\documentclass[runningheads]{llncs}
\usepackage[T1]{fontenc}
\usepackage{array}
\usepackage{booktabs}
\usepackage{multirow}
\usepackage[table,xcdraw]{xcolor}
\usepackage{marvosym}
\usepackage{graphicx}
\usepackage{amsmath}
\usepackage[ruled]{algorithm2e}
\usepackage[colorlinks,linkcolor=red]{hyperref}
\makeatletter
\def\thanks{\protected@xdef\@thanks{\@thanks\protect\footnotetext{\dag\ Equal contributions}}}
\makeatother

\newcommand{\ie}{{\it i.e.,}}

\begin{document}
\title{WeakPolyp: You Only Look Bounding Box for Polyp Segmentation}

%%%%%%%%%%%%%%%%%%%%%%%%%%%%%%%%%%%%%%%%%%%%%%%%%%%%%  
%%
%% authors
%% 
%%%%%%%%%%%%%%%%%%%%%%%%%%%%%%%%%%%%%%%%%%%%%%%%%%%%% 
% \author{Anonymous}
% \institute{Paper ID: 147}

\author{
    Jun Wei\inst{1,2,\dag},
    Yiwen Hu\inst{1,2,5,\thanks{\dag}},
    Shuguang Cui\inst{2,1},
    S.Kevin Zhou\inst{3,4},     
    Zhen Li\inst{2,1,\textrm{\Letter}} 
}% index{Wei Jun, Hu Yiwen, Cui Shuguang, Zhou S.Kevin, Li Zhen}

\institute{
    $^1$ FNii, CUHK-Shenzhen, Shenzhen, China
    $^2$ SSE, CUHK-Shenzhen, Shenzhen, China\\
    $^3$ School of Biomedical Engineering \& Suzhou Institute for Advanced Research, University of Science and Technology of China, Suzhou, China\\
    $^4$ Institute of Computing Technology, Chinese Academy of Sciences, Beijing, China\\
    $^5$ South China Hospital, Shenzhen University, Shenzhen, China\\
    \email{\{junwei@link., lizhen@\}cuhk.edu.cn}
}

\maketitle

%%%%%%%%%%%%%%%%%%%%%%%%%%%%%%%%%%%%%%%%%%%%%%%%%%%%%  
%%
%% abstract
%% 
%%%%%%%%%%%%%%%%%%%%%%%%%%%%%%%%%%%%%%%%%%%%%%%%%%%%% 
\begin{abstract}
Limited by expensive pixel-level labels, polyp segmentation models are plagued by data shortage and suffer from impaired generalization. In contrast, polyp bounding box annotations are much cheaper and more accessible. 
Thus, to reduce labeling cost, we propose to learn a weakly supervised polyp segmentation model (\ie WeakPolyp) completely based on bounding box annotations.
However, coarse bounding boxes contain too much noise. To avoid interference, we introduce the mask-to-box (M2B) transformation. By supervising the outer box mask of the prediction instead of the prediction itself, M2B greatly mitigates the mismatch between the coarse label and the precise prediction.
But, M2B only provides sparse supervision, leading to non-unique predictions.
Therefore, we further propose a scale consistency (SC) loss for dense supervision. By explicitly aligning predictions across the same image at different scales, the SC loss largely reduces the variation of predictions. 
Note that our WeakPolyp is a plug-and-play model, which can be easily ported to other appealing backbones. Besides, the proposed modules are only used during training, bringing no computation cost to inference. 
Extensive experiments demonstrate the effectiveness of our proposed WeakPolyp, which surprisingly achieves a comparable performance with a fully supervised model, requiring no mask annotations at all. Codes are available at \url{https://github.com/weijun88/WeakPolyp}.

\keywords{Polyp segmentation \and Weak Supervision \and Colorectal cancer}
\end{abstract}

%%%%%%%%%%%%%%%%%%%%%%%%%%%%%%%%%%%%%%%%%%%%%%%%%%%%%  
%%
%% introduction
%% 
%%%%%%%%%%%%%%%%%%%%%%%%%%%%%%%%%%%%%%%%%%%%%%%%%%%%% 
\section{Introduction}
% - 背景介绍
%     - 结直肠癌的高发性
%     - 肠镜早期筛查可以有效预防结肠癌
%     - 自动化息肉分割可以辅助肠镜筛查，防止漏检
% - 息肉分割的难点
%     - 大规模像素级标注成本高
%     - 像素级标注存在很多误差
Colorectal Cancer (CRC) has become a major threat to health worldwide. Since most CRCs originate from colorectal polyps, early screening for polyps is necessary. Given its significance, automatic polyp segmentation models~\cite{PraNet,PNSNet,wang2022stepwise,BoxPolyp} have been designed to aid in screening. For example, ACSNet~\cite{zhang2020adaptive}, HRENet~\cite{shen2021hrenet}, LDNet~\cite{LDNet} and CCBANet~\cite{nguyen2021ccbanet} propose to use convolutional neural networks to extract multi-scale contexts for robust predictions. LODNet~\cite{cheng2021learnable}, PraNet~\cite{PraNet}, and MSNet~\cite{zhao2021automatic} aim to improve the model's discrimination of polyp boundaries. SANet~\cite{SANet} eliminates the distribution gap between the training set and the testing set, thus improving the model generalization. Recently, TGANet~\cite{TGANet} introduces text embeddings to enhance the model's discrimination. Furthermore, Transfuse~\cite{Transfuse}, PPFormer~\cite{cai2022using}, and Polyp-Pvt~\cite{Polyp-PVT} introduce the Transformer~\cite{dosovitskiy2020vit} backbone to extract global contexts, achieving a significant performance gain.

\begin{figure}[t]
  \centering
  \includegraphics[scale=0.45]{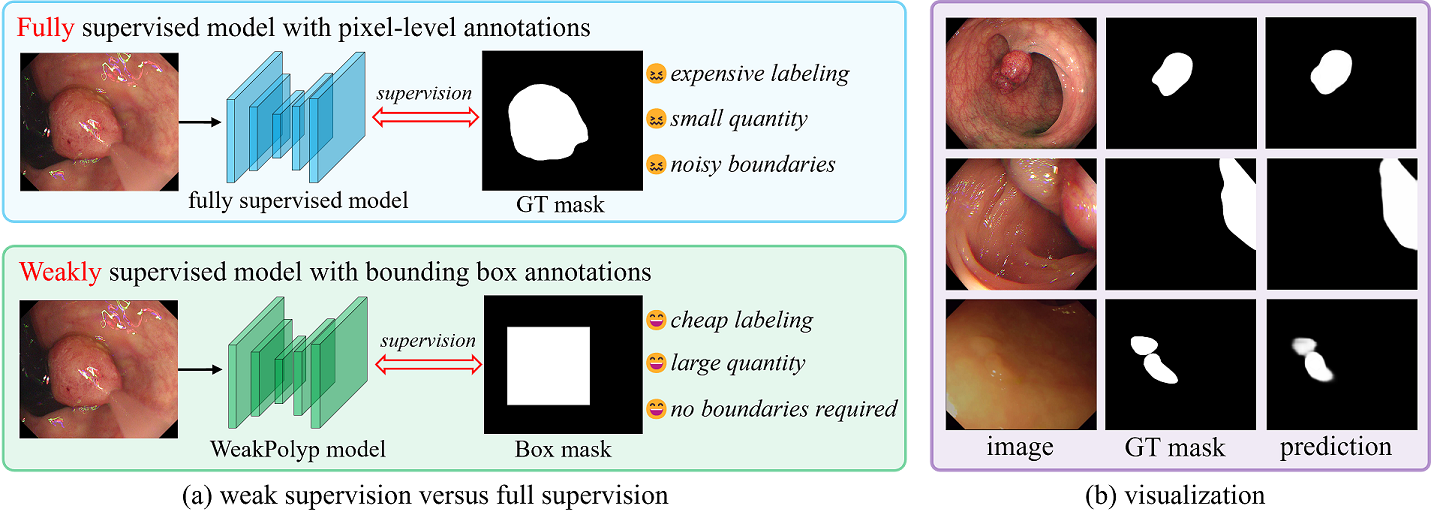}
  \caption{(a) Comparison between the fully supervised model and our proposed WeakPolyp using box mask only. (b) Visualization of prediction from WeakPolyp.}
  \label{fig:teaser}
\end{figure}

% - 解决方法
%     - 使用box级别的标注，即可实现弱监督的息肉分割
%     - 只需要标注两个点，极大的降低标注成本
%     - 在相同标注量下，可以标注更多的图像，在大规模息肉分割上更适用
%     - 对噪声容忍度更好，不要求精确的边缘分割
%     - 达到了和全监督相近的性能
All above models are fully supervised and require pixel-level annotations. However, pixel-by-pixel labeling is time-consuming and expensive, which hampers practical clinical usage. Besides, many polyps do not have well-defined boundaries. Pixel-level labeling inevitably introduces subjective noise. To address the above limitations, a generalized polyp segmentation model is urgently needed. In this paper, we achieve this goal by a weakly supervised polyp segmentation model (named \textbf{WeakPolyp}) that only uses coarse bounding box annotations. Fig.~\ref{fig:teaser}(a) shows the differences between our WeakPolyp and fully supervised models. Compared with fully supervised ones, WeakPolyp requires only a bounding box for each polyp, thus dramatically reducing the labeling cost. More meaningfully, WeakPolyp can take existing large-scale polyp detection datasets to assist the polyp segmentation task. Finally, WeakPolyp does not require the labeling for polyp boundaries, avoiding the subjective noise at source. All these advantages make WeakPolyp more clinically practical.

% - box级别监督的难点
%     - 粗糙，不好直接利用   
%     - 稀疏，只有部分像素参与监督
%     - 取得了和全监督相匹配的性能
However, bounding box annotations are much coarser than pixel-level ones, which can not describe the shape of polyps. Simply adopting these box annotations as supervision introduces too much background noise, thereby leading to suboptimal models. As a solution, BoxPolyp~\cite{BoxPolyp} only supervises the pixels with high certainty. However, it requires a fully supervised model to predict the uncertainty map. Unlike BoxPolyp, our WeakPolyp completely follows the weakly supervised form that requires no additional models or annotations. Surprisingly, just by redesigning the supervision loss without any changes to the model structure, WeakPolyp achieves comparable performance to its fully supervised counterpart. Fig.~\ref{fig:teaser}(b) visualizes some predicted results by WeakPolyp.

% - 我们所用方法的组成部分
%     - 改进了box的监督形式
%     - 为无标注区域引入了像素级的自监督学习
WeakPolyp is mainly enabled by two novel components: mask-to-box (M2B) transformation and scale consistency (SC) loss. In practice, M2B is applied to transform the predicted mask into a box-like mask by projection and back-projection. Then, this transformed mask is supervised by the bounding box annotation. This indirect supervision avoids the misleading of box-shape bias of annotations. However, many regions in the predicted mask are lost in the projection and therefore get no supervision. To fully explore these regions, we propose the SC loss to provide a pixel-level self-supervision while requiring no annotations at all. Specifically, the SC loss explicitly reduces the distance between predictions of the same image at different scales. By forcing feature alignment, it inhibits the excessive diversity of predictions, thus improving the model generalization.

% - 贡献总结
%     - 完全使用box级别标注实现了像素级预测
%     - 在不更改网络结构的基础上，提出了两个监督损失，并达到了和全监督相匹配的性能
%     - 我们在不同的数据集和网络结构上验证了所提方法的有效性
In summary, our contributions are three-fold: (1) We build the WeakPolyp model completely based on bounding box annotations, which largely reduces the labeling cost and achieves a comparable performance to full supervision. (2) We propose the M2B transformation to mitigate the mismatch between the prediction and the supervision, and design the SC loss to improve the robustness of the model against the variability of the predictions. (3) Our proposed WeakPolyp is a plug-and-play option, which can boost the performances of polyp segmentation models under different backbones.

%%%%%%%%%%%%%%%%%%%%%%%%%%%%%%%%%%%%%%%%%%%%%%%%%%%%%  
%%
%% methods
%% 
%%%%%%%%%%%%%%%%%%%%%%%%%%%%%%%%%%%%%%%%%%%%%%%%%%%%% 
\section{Method}
\begin{figure}[t]
  \centering
  \includegraphics[scale=0.6]{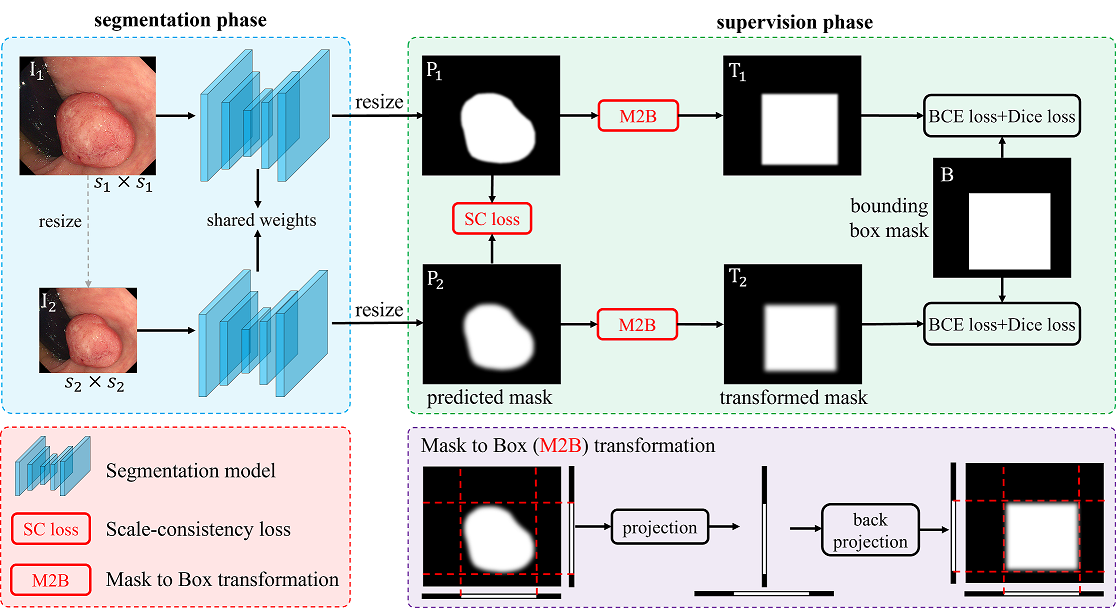}
  \caption{The framework of our proposed WeakPolyp model, which consists of the segmentation phase and the supervision phase. The segmentation phase predicts the polyp mask for each input firstly, and the supervision phase uses the coarse box annotation to guide previous predicted mask. Note that our contributions mainly lie in the supervision phase, where the proposed M2B transformation converts the predicted mask into a box mask to accommodate the bounding box annotation. Besides, another proposed SC loss is introduced to provide dense supervision from multi-scales, which improves the consistency of predictions.}
  \label{fig:framework}
\end{figure}

% \subsection{Model Introduction}
\noindent\textbf{Model Components.}
Fig.~\ref{fig:framework} depicts the components of WeakPolyp, including the segmentation phase and the supervision phase. For the segmentation phase, we adopt Res2Net~\cite{GaoCZZYT21} as the backbone. For input image $I\in R^{H \times W}$, Res2Net extracts four scales of features $\{f_i | i=1,...,4 \}$ with the resolutions $[\frac{H}{2^{i+1}}, \frac{W}{2^{i+1}}]$. Considering the computational cost, only $f_2, f_3$ and $f_4$ are utilized. To fuse them, we first apply a $1\times 1$ convolutional layer to unify the channels of $f_2,f_3,f_4$ and then use the bilinear upsampling to unify their resolutions. After being transformed to the same size, $f_2,f_3,f_4$ are added together and fed into one $1\times 1$ convolutional layer for final prediction. Instead of the segmentation phase, our contributions primarily lie in the supervision phase, including mask-to-box (M2B) transformation and scale consistency (SC) loss. Notably, both M2B and SC are independent of the specific model structure.

\noindent\textbf{Model Pipeline.}
For each input image $I$, we first resize it into two different scales: $I_1 \in R^{s_1 \times s_1}$ and $I_2 \in R^{s_2 \times s_2}$. Then, $I_1$ and $I_2$ are sent to the segmentation model and get two predicted masks $P_1$ and $P_2$, both of which have been resized to the same size. Next, an SC loss is proposed to reduce the distance between $P_1$ and $P_2$, which helps suppress the variation of the prediction. Finally, to fit the bounding box annotations ($B$), $P_1$ and $P_2$ are sent to M2B and converted into box-like masks $T_1$ and $T_2$. With $T_1/T_2$ and $B$, we calculate the binary cross entropy (BCE) loss and Dice loss, without worrying about noise interference.

\subsection{Mask-to-Box (M2B) Transformation}
%\textbf{Introduction.} 
% - 直接使用 box mask 做像素级监督，会混入很多背景噪声，导致性能退化
% - 求预测 mask 的外接矩形，再用 box 标签监督它，不用担心噪声干扰
% - 实现步骤：投影，反投影
One naive method to achieve the weakly supervised polyp segmentation is to use the bounding box annotation $B$ to supervise the predicted mask $P_1/P_2$. Unfortunately, models trained in this way show poor generalization. Because there is a strong box-shape bias in $B$. Training with this bias, the model is forced to predict the box-shape mask, unable to maintain the polyp's contours. To solve this, we innovatively use $B$ to supervise the bounding box mask (\ie $T_1/T_2$) of $P_1/P_2$, rather than $P_1/P_2$ itself. This indirect supervision separates $P_1/P_2$ from $B$ so that $P_1/P_2$ is not affected by the shape bias of $B$ while obtaining the position and extent of polyps. But how to implement the transformation from $P_1/P_2$ to $T_1/T_2$? We design the M2B module, which consists of two steps: projection and back-projection, as shown in Fig.~\ref{fig:framework}.

\noindent\textbf{Projection.}
As shown in Eq.~\ref{eq:projection}, given a predicted mask $P\in [0,1]^{H\times W}$, we project it horizontally and vertically into two vectors $P_w \in [0,1]^{1 \times W}$ and $P_h \in [0,1]^{H\times 1}$. In this projection, instead of using mean pooling, we use max pooling to pick the maximum value for each row/column in $P$. Because max pooling can completely remove the shape information of the polyp. After projection, only the position and scope of the polyp are stored in $P_w$ and $P_h$.
\begin{equation}
  \begin{aligned}
      P_w = \max(P, \text{axis}=0) \in [0,1]^{1 \times W}, \quad
      P_h = \max(P, \text{axis}=1) \in [0,1]^{H \times 1} 
      \label{eq:projection}
  \end{aligned}
\end{equation}

\noindent\textbf{Back-projection.}
% 将 p_w 和 p_h 先扩增到相同的形状
% 求二者的极小值
Based on $P_w$ and $P_h$, we construct the bounding box mask of the polyp by back-projection. As shown in Eq.~\ref{eq:backprojection}, $P_w$ and $P_h$ are first repeated into $P_w^{'}$ and $P_h^{'}$ with the same size as $P$. Then, we element-wisely take the minimum of $P_w^{'}$ and $P_h^{'}$ to achieve the bounding box mask $T$. As shown in Fig.~\ref{fig:framework}, $T$ no longer contains the contours of the polyp.
\begin{equation}
  \begin{aligned}
      P_w^{'}&= \text{repeat}(P_w, H, \text{axis}=0) \in [0,1]^{H \times W}\\
      P_h^{'}&= \text{repeat}(P_h, W, \text{axis}=1) \in [0,1]^{H \times W}\\
      T &= \min(P_w^{'}, P_h^{'}) \in [0,1]^{H \times W}
      \label{eq:backprojection}
  \end{aligned}
\end{equation}

\noindent\textbf{Supervision.}
By M2B, $P_1$ and $P_2$ are transformed into $T_1$ and $T_2$, respectively. Because both $T_1/T_2$ and $B$ are box-like masks, we directly calculate the supervision loss between them without worrying about the misguidance of box-shape bias. Specifically, we follow~\cite{SANet,PraNet} to adopt BCE loss $\mathcal{L}_{BCE}$ and Dice loss $\mathcal{L}_{Dice}$ for model supervision, as shown in Eq.~\ref{eq:loss:BD}.
\begin{equation}
  \begin{aligned}
    \mathcal{L}_{Sum} = \frac{\mathcal{L}_{BCE}(T_1, B)+\mathcal{L}_{BCE}(T_2, B)}{2} + \frac{\mathcal{L}_{Dice}(T_1, B)+\mathcal{L}_{Dice}(T_2, B)}{2}
    \label{eq:loss:BD}
  \end{aligned}
\end{equation}

\noindent\textbf{Priority.}
% 相较于直接监督P，我们设计M2B避免了噪声
% 可导，可并行，实现简单
By simple transformation, M2B turns the noisy supervision into a noise-free one, so that the predicted mask is able to preserve the contours of the polyp. Notably, M2B is differentiable, which can be easily implemented with PyTorch and plugged into the model to participate in gradient backpropagation.

\subsection{Scale Consistency (SC) Loss}
% 动机
%   - 只有部分像素有监督
%   - 和全监督的分割在监督上有差异，缺失像素级的标注
%   - 预测结果具有歧义性，存在多对一的情况
%   - 没有考虑图像本身的信息
%   - 需要引入更多监督
% 方法特性
%   - 只是在loss层面进行了一些变换，没有对网络进行任何修改
%   - 具有通用性，可以被迁移不同的方法上

\begin{figure}[t]
  \centering
  \includegraphics[scale=0.52]{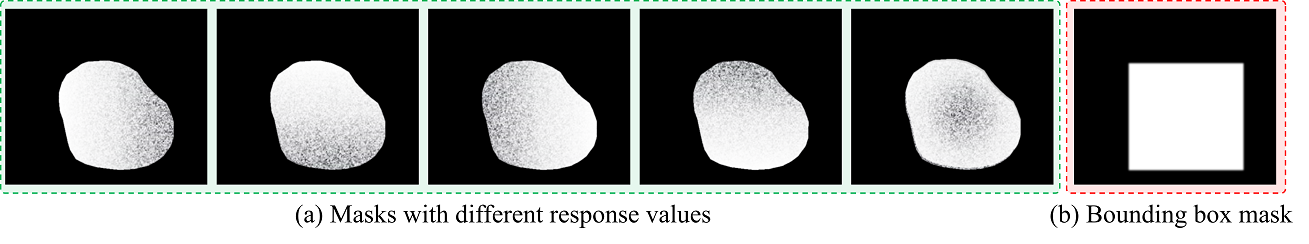}
  \caption{Different predictions may correspond to the same bounding box mask.}
  \label{fig:inconsistency}
\end{figure}

%\noindent\textbf{Introduction.}
In M2B, most pixels in $P$ are ignored in the projection, thus only a few pixels with high response values are involved in the supervision loss. This sparse supervision may lead to non-unique predictions. As shown in Fig.~\ref{fig:inconsistency}, after M2B projection, five predicted masks with different response values can be transformed into the same bounding box mask. Therefore, we consider introducing the SC loss to achieve dense supervision without annotations, which reduces the degree of freedom of predictions.

\noindent\textbf{Method.} 
As shown in Fig.~\ref{fig:framework}, due to the non-uniqueness of the prediction and the scale difference between $I_1$ and $I_2$, $P_1$ and $P_2$ differ in response values. But $P_1$ and $P_2$ come from the same image $I_1$. They should be exactly the same. Given this, as shown in Eq.~\ref{eq:loss:SC}, we build the dense supervision $\mathcal{L}_{SC}$ by explicitly reducing the distance between $P_1$ and $P_2$, where $(i,j)$ is the pixel coordinates. Note that only pixels inside bounding box are involved in $\mathcal{L}_{SC}$ to emphasize more on polyp regions. Despite its simplicity, $\mathcal{L}_{SC}$ brings pixel-level constraints to compensate for the sparsity of $\mathcal{L}_{Sum}$, thus reducing the variety of predictions.
\begin{equation}
  \begin{aligned}
    \mathcal{L}_{SC} = \frac{\sum_{(i,j)\in box}|P_1^{i,j}-P_2^{i,j}|}{\sum_{(i,j)\in box}{1}} \label{eq:loss:SC}\\
    \end{aligned}
\end{equation}

\subsection{Total Loss}
As shown in Eq.~\ref{eq:loss:Total}, combining $\mathcal{L}_{Sum}$ and $\mathcal{L}_{SC}$ together, we get WeakPolyp model. Note that WeakPolyp simply replaces the supervision loss without making any changes to the model structure. Therefore, it is general and can be ported to other models. Besides, $\mathcal{L}_{Sum}$ and $\mathcal{L}_{SC}$ are only used during training. In inference, they will be removed, thus having no effect on the speed of the model.
\begin{equation}
  \begin{aligned}
    \mathcal{L}_{Total} = \mathcal{L}_{Sum}+\mathcal{L}_{SC} \label{eq:loss:Total}
    \end{aligned}
\end{equation}

\begin{table*}[!t]
  \centering
  \setlength\tabcolsep{4.0pt}
  \caption{Quantitative comparison between different baselines and our WeakPolyp, involving two datasets (SUN-SEG and POLYP-SEG) and two backbones (Res2Net-50~\cite{GaoCZZYT21} and PVTv2-B2~\cite{wang2021pvtv2}). The \textbf{gt} row is the performance upper bound. The \textbf{box} row is the performance lower bound. \textbf{'Bac.'} means backbone. \textbf{'Sup.'} means supervision. The highest and second-highest scores are marked in \textcolor{red}{red} and \textcolor{blue}{blue}, respectively}
  \label{tab:baseline}
  
  \begin{tabular}{c|c|cccccccccc}
      \hline\toprule
      \multirow{3}{*}{\textbf{Bac.}} & \multirow{3}{*}{\textbf{Sup.}}
      & \multicolumn{6}{c}{\textbf{SUN-SEG}} & \multicolumn{4}{c}{\textbf{POLYP-SEG}} \\  
      \cmidrule(l){3-8}\cmidrule(l){9-12}
      &  & \multicolumn{2}{c}{Easy Testing} & \multicolumn{2}{c}{Hard Testing} & \multicolumn{2}{c}{Training} & \multicolumn{2}{c}{Testing} & \multicolumn{2}{c}{Training} \\
      \cmidrule(l){3-4}\cmidrule(l){5-6}\cmidrule(l){7-8}\cmidrule(l){9-10}\cmidrule(l){11-12}
      &  & Dice & IoU & Dice & IoU & Dice & IoU & Dice & IoU & Dice & IoU \\
      
      \hline\toprule
      \multirow{4}{*}{Res.}
      & gt      & \textcolor{blue}{.772} & \textcolor{blue}{.693} & \textcolor{blue}{.798} & \textcolor{blue}{.716} & \textcolor{red}{.931} & \textcolor{red}{.876} & \textcolor{red}{.761} & \textcolor{red}{.684} & \textcolor{red}{.936} & \textcolor{red}{.884}  \\
      & grabcut & .595 & .514 & .617 & .530 & .706 & .608 & .660 & .579 & .778 & .687  \\
      & box     & .715 & .601 & .718 & .599 & .806 & .685 & .686 & .566 & .804 & .683  \\
      \rowcolor{black!10}
      & \textbf{Ours} & \textcolor{red}{.792} & \textcolor{red}{.715} & \textcolor{red}{.807} & \textcolor{red}{.727} & \textcolor{blue}{.899} & \textcolor{blue}{.826} & \textcolor{blue}{.760} & \textcolor{blue}{.680} & \textcolor{blue}{.909} & \textcolor{blue}{.842} \\

      \hline\toprule
      \multirow{4}{*}{Pvt.}
      & gt      & \textcolor{blue}{.851} & \textcolor{blue}{.780} & \textcolor{red}{.858} & \textcolor{red}{.784} & \textcolor{red}{.932} & \textcolor{red}{.878} & \textcolor{red}{.793} & \textcolor{red}{.715} & \textcolor{red}{.936} & \textcolor{red}{.883}  \\
      & grabcut & .741 & .648 & .747 & .649 & .766 & .670 & .644 & .559 & .780 & .683  \\
      & box     & .769 & .652 & .770 & .648 & .804 & .681 & .734 & .611 & .824 & .705  \\
      \rowcolor{black!10}
      & \textbf{Ours} & \textcolor{red}{.853} & \textcolor{red}{.781} & \textcolor{blue}{.854} & \textcolor{blue}{.777} & \textcolor{blue}{.907} & \textcolor{blue}{.839} & \textcolor{blue}{.792} & \textcolor{blue}{.707} & \textcolor{blue}{.922} & \textcolor{blue}{.859}  \\
      \hline\toprule
  \end{tabular}
\end{table*}

%%%%%%%%%%%%%%%%%%%%%%%%%%%%%%%%%%%%%%%%%%%%%%%%%%%%%  
%%
%% Experiments
%% 
%%%%%%%%%%%%%%%%%%%%%%%%%%%%%%%%%%%%%%%%%%%%%%%%%%%%% 
\section{Experiments}
\noindent\textbf{Datasets.} 
Two large polyp datasets are adopted to evaluate the model performance, including SUN-SEG~\cite{PNS+} and POLYP-SEG. SUN-SEG originates from~\cite{SUN-Data,SUN-Dev}, which consists of 19,544 training images, 17,070 easy tesing images, and 12,522 hard testing images. POLYP-SEG is our private polyp segmentation dataset, which contains 15,916 training images and 4,040 testing images. Note that, during training, only bounding box annotations are adopted in our WeakPolyp. 

\noindent\textbf{Training Settings.}
WeakPolyp is implemented using PyTorch. All input images are uniformly resized to 352×352. For data augmentation, random flip, random rotation, and multi-scale training are adopted. The whole network is trained in an end-to-end way with an AdamW optimizer. Initial learning rate and batch size are set to 1e-4 and 16, respectively. We train the entire model for 16 epochs.

\noindent\textbf{Quantitative Comparison.}
Table.~\ref{tab:baseline} compares the model performance under different supervisions, backbones, and datasets. The overall performance order is \textit{gt>WeakPolyp>box>grabcut}. The model supervised by grabcut~\cite{rother2004grabcut} masks performs the worst, because the foreground and background of polyp images are similar. Grabcut can not well distinguish between them, resulting in poor masks. Our WeakPolyp predictably outperforms the model supervised by box masks because it is not affected by the box-shape bias of the annotations. Interestingly, WeakPolyp even surpasses the fully supervised model on SUN-SEG, which indicates that there is a lot of noise in the pixel-level annotations. But WeakPolyp does not require pixel-level annotations so it avoids noise interference.

\begin{figure*}[!t]
  \centering
  \includegraphics[scale=0.36]{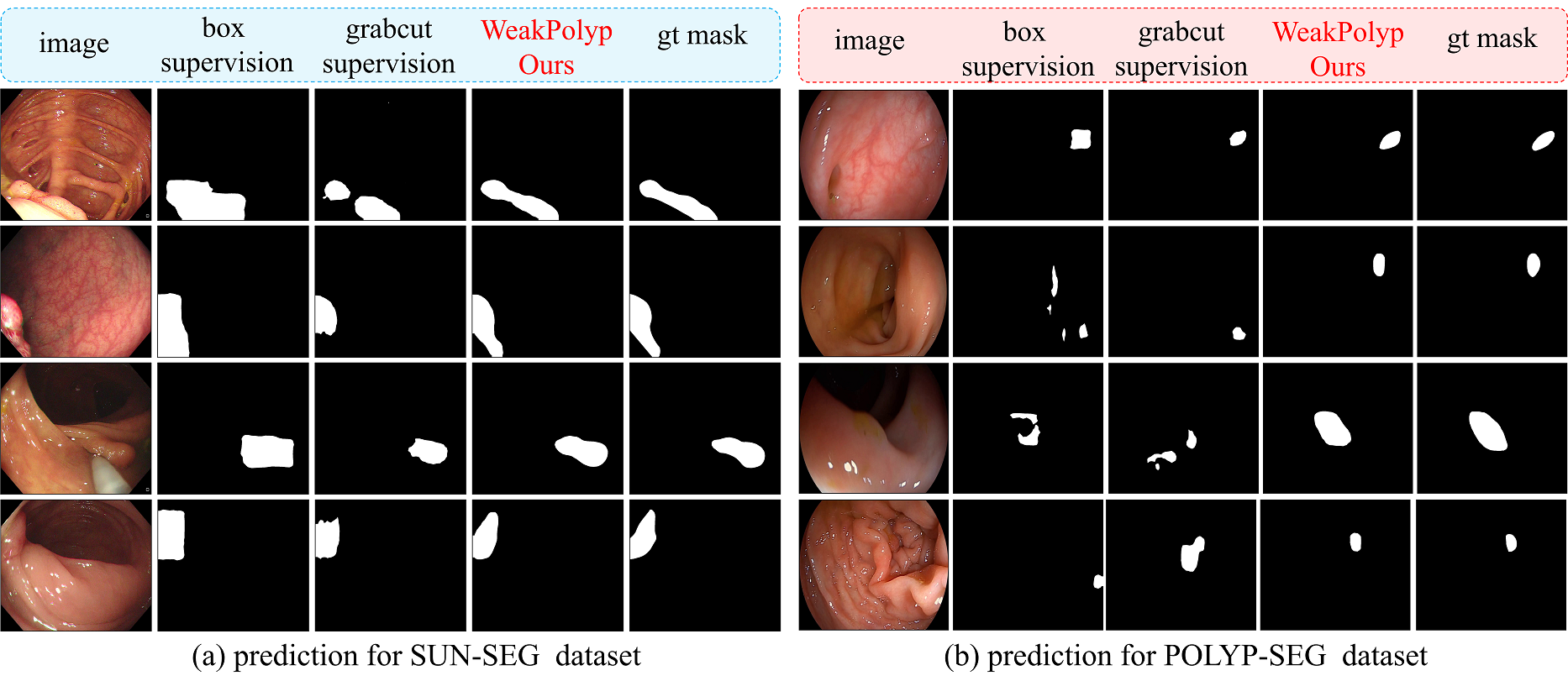}
  \caption{Visualization comparison between predictions based on different supervisions.}
  \label{fig:visualization}
\end{figure*}

\begin{table*}[!t]
  \centering
  \setlength\tabcolsep{6.5pt}
  \caption{Ablation studies on the SUN-SEG testing set under different backbones.}
  \label{tab:ablation}
  
  \begin{tabular}{l|cccccccccc}
      \hline\toprule
      \multirow{3}{*}{\textbf{Modules}}
      & \multicolumn{4}{c}{\textbf{Res2Net-50}} & \multicolumn{4}{c}{\textbf{PVTv2-B2}} \\  
      \cmidrule(l){2-5}\cmidrule(l){6-9}
      &  \multicolumn{2}{c}{Easy Testing} & \multicolumn{2}{c}{Hard Testing} & \multicolumn{2}{c}{Easy Testing} & \multicolumn{2}{c}{Hard Testing}\\
      \cmidrule(l){2-3}\cmidrule(l){4-5}\cmidrule(l){6-7}\cmidrule(l){8-9}
      & Dice & IoU & Dice & IoU & Dice & IoU & Dice & IoU \\
      
      \hline\toprule
      Base         & .715 & .601 & .718 & .599 & .769 & .652 & .770 & .648\\
      Base+M2B     & .748 & .654 & .768 & .673 & .822 & .738 & .822 & .735\\
      Base+M2B+SC  & .792 & .715 & .807 & .727 & .853 & .781 & .854 & .777 \\
      \hline\toprule
  \end{tabular}
\end{table*}

\noindent\textbf{Visual Comparison.} 
Fig.~\ref{fig:visualization} visualizes some predictions based on different supervisions. Compared with other counterparts, WeakPolyp not only highlights the polyp shapes but also suppresses the background noise. Even for challenging scenarios, WeakPolyp still handles well and generates accurate masks.

\noindent\textbf{Ablation Study.} 
To investigate the importance of each component in WeakPolyp, we evaluate the model on both Res2Net-50 and PVTv2-B2 for ablation studies. As shown in Table~\ref{tab:ablation}, all proposed modules are beneficial for the final predictions. Combining all these modules, our model achieves the highest performance.

\noindent\textbf{Compared with Fully Supervised Methods.} 
Table.~\ref{tab:fullsupervision} shows our WeakPolyp is even superior to many previous fully supervised methods: PraNet~\cite{PraNet}, SANet~\cite{SANet}, 2/3D~\cite{2D/3D} and PNS+~\cite{PNS+}, which shows the excellent application prospect of weakly supervised learning in the polyp field.

% \noindent\textbf{Extension.}
% Not limited to weakly supervised learning, WeakPolyp can also be integrated into the fully supervised models to take advantage of both pixel-level annotations and bounding box ones, which greatly expands the range of data available to a single model. Specifically, we port WeakPolyp into SANet and Polyp-Pvt, and adopt five datasets for evaluation: CVC-ColonDB~\cite{bernal2012towards}, Kvasir~\cite{jha2020kvasir}, CVC-ClinicDB~\cite{bernal2015wm}, EndoScene~\cite{vazquez2017benchmark} and ETIS~\cite{silva2014toward}. We follow the same dataset partition as~\cite{PraNet} but add the SUN-SEG training set with bounding box annotations as an extra. As shown in Table.~\ref{tab:extension}, WeakPolyp can greatly boost the fully supervised models under different backbones.

\begin{table*}[!t]
  \centering
  \setlength\tabcolsep{7pt}
  \caption{Performance comparison with previous fully supervised models on SUN-SEG.}
  \label{tab:fullsupervision}
  \begin{tabular}{l|l|l|cccc} 
        \hline \toprule
        \multirow{2}{*}{\textbf{Model}} & \multirow{2}{*}{\textbf{Conference}} &\multirow{2}{*}{\textbf{Backbone}} & \multicolumn{2}{c}{\textbf{Easy Testing}} & \multicolumn{2}{c}{\textbf{Hard Testing}} \\ 
        \cmidrule(l){4-5} \cmidrule(l){6-7}
        &  & & Dice & IoU & Dice & IoU \\ 
         \hline \toprule
        PraNet~\cite{PraNet}    & MICCAI 2020  & Res2Net-50   & .689 & .608 & .660 & .569 \\
        2/3D~\cite{2D/3D}       & MICCAI 2020  & ResNet-101   & .755 & .668 & .737 & .643 \\
        SANet~\cite{SANet}      & MICCAI 2021  & Res2Net-50   & .693 & .595 & .640 & .543 \\
        PNS+~\cite{PNS+}        & MIR 2022     & Res2Net-50   & .787 & .704 & .770 & .679 \\
        \hline
        \rowcolor{black!10}
        \textbf{Ours} & & Res2Net-50 & \textbf{.792}  & \textbf{.715}  & \textbf{.807}  & \textbf{.727}  \\
        \rowcolor{black!10}
        \textbf{Ours} & & PVTv2-B2 & \textbf{.853}  & \textbf{.781}  & \textbf{.854}  & \textbf{.777}  \\
        \hline \toprule
  \end{tabular}
\end{table*}

\section{Conclusion}
Limited by expensive labeling cost, pixel-level annotations are not readily available, which hinders the development of the polyp segmentation field. In this paper, we propose the WeakPolyp model completely based on bounding box annotations. WeakPolyp requires no pixel-level annotations, thus avoiding the interference of subjective noise labels. More importantly, WeakPolyp even achieves a comparable performance to the fully supervised models, showing the great potential of weakly supervised learning in the polyp segmentation field. In future, we will introduce temporal information into weakly supervised polyp segmentation to further reduce the model's dependence on labeling.

\section{Acknowledgement}
% This work is supported by the Guangdong Provincial Key Laboratory of Big Data Computing, The Chinese University of Hong Kong, Shenzhen, by NSFC-Youth 61902335, by Key Area R\&D Program of Guangdong Province with
% grant No.2018B030338001, by the National Key R\&D Program of China with grant No.2018YFB1800800, by Shenzhen Outstanding Talents Training Fund, by Guangdong Research Project No.2017ZT07X152, by Guangdong Regional Joint Fund-Key Projects 2019B1515120039, by the NSFC 61931024\&81922046, by helixon biotechnology company Fund and CCF-Tencent Open Fund.

This work was supported in part by Shenzhen General Program No.\\JCYJ20220530143600001, by the Basic Research Project No. HZQB-KCZYZ-2021067 of Hetao Shenzhen HK S\&T Cooperation Zone, by Shenzhen-Hong Kong Joint Funding No. SGDX20211123112401002, by Shenzhen Outstanding Talents Training Fund, by Guangdong Research Project No. 2017ZT07X152 and No. 2019CX01X104, by the Guangdong Provincial Key Laboratory of Future Networks of Intelligence (Grant No. 2022B1212010001), by the Guangdong Provincial Key Laboratory of Big Data Computing, The Chinese University of Hong Kong, Shenzhen, by the NSFC 61931024\&81922046, by zelixir biotechnology company Fund, by Tencent Open Fund.

\bibliographystyle{bibstyle}
\bibliography{bibliography}
\end{document}